\documentclass[11pt]{article}

\usepackage[preprint]{acl}

\usepackage{times}
\usepackage{latexsym}

\usepackage[T1]{fontenc}

\usepackage[utf8]{inputenc}

\usepackage{microtype}

\usepackage{inconsolata}

\usepackage{graphicx}

\usepackage{kotex}
\usepackage{booktabs}
\usepackage{multirow}
\usepackage{placeins}
\usepackage{colortbl}
\newcommand{\oursrow}{\rowcolor{gray!12}} 
\usepackage{hyperref}

\usepackage[utf8]{inputenc}
\usepackage{array}
\usepackage{makecell}
\usepackage{subcaption}
\usepackage{amssymb}
\usepackage{amsmath}
\usepackage{cleveref}
\usepackage{listings}
\usepackage{tabularx}
\newcommand{\eg}{\emph{e.g.}}

\usepackage{amsmath,amsfonts,bm}
\newcommand{\ie}{\textit{i}.\textit{e}., }

\title{Who Should Lead Decoding Now? Tracking Reliable Trajectories for Ensembling Masked Diffusion Language Models}

\author{
Heecheol Yun$^{1}$ \quad
Joowon Kim$^{1}$ \quad
Eunho Yang$^{1,2}$ \quad
Joonhyung Park$^{1}$ \\
$^{1}$KAIST \quad
$^{2}$AITRICS \\
{\tt\small \{yoon6503, deepjoon, kjwispro, eunhoy\}@kaist.ac.kr} \\
}

\begin{document}
\maketitle

\begin{abstract}
Masked Diffusion Language Models (MDLMs) have emerged as a distinct paradigm for sequence generation.
As MDLMs become diverse in capabilities and knowledge coverage, an important question is how to combine their knowledge.
Toward this, we first investigate the unique decoding dynamics of MDLMs. We find that successful generations exhibit stable confidence dynamics over answer-relevant positions, while unreliable trajectories can often be corrected by injecting promising intermediate states from other models. 
Guided by this observation, we propose \textbf{TIE} (\textbf{T}rajectory-based \textbf{I}terative \textbf{E}nsembling), a knowledge fusion framework in which MDLMs iteratively identify reliable decoding trajectories and relay them across models.
TIE tracks confidence dynamics over answer-relevant positions to determine which model currently follows a more reliable trajectory and selectively transfers partially denoised sequences across models. 
As the model on the more promising trajectory often changes across denoising steps, TIE allows different models to contribute complementary strengths at different stages of generation.
Strong performance across diverse reasoning tasks, along with our analyses, suggests that TIE offers a practical approach to the underexplored problem of MDLM ensembling. 
\end{abstract}

\section{Introduction}
Masked Diffusion Language Models (MDLMs) are becoming increasingly compelling alternatives to the autoregressive paradigm. By denoising masked sequences in parallel through an iterative remasking process, MDLMs show competitive sequence generation capabilities across a broad range of domains~\citep{sahoo2024simple,dream,llada,ye2025beyond}.

As the family of MDLMs continues to diversify with models exhibiting different strengths, training distributions, and decoding dynamics, one question naturally comes up: \textit{how can we effectively orchestrate or fuse knowledge from heterogeneous MDLMs?} This question has grown important in recent years, as users often try different models jointly on their own tasks in search of the best possible results. However, such ensembling strategies for MDLMs remain largely underexplored.

One straightforward approach would be to extend conventional ensemble approaches in autoregressive language models: taking into account the next-token probability distributions, then averaging them~\citep{yu-etal-2024-breaking,xu-etal-2024-bridging} or routing toward the more confident model~\citep{shen-etal-2024-learning,wang2025speculate}. These approaches, however, are not directly applicable to MDLMs due to their unique decoding dynamics. Since sequences are generated in a flexible, non-left-to-right order, each model may operate on different partially denoised sequences at each step, which makes it difficult to define a shared next-token across models. Such disparities call for knowledge orchestration frameworks specifically designed for MDLMs.

Toward this, we first scrutinize the decoding dynamics of MDLMs to gain insights that guide the design of a knowledge ensemble framework for generating quality-enhanced responses. Specifically, we focus on two perspectives: (i) identifying the more confident model that is likely to produce a correct answer before the full response is generated, allowing it to lead the ensemble process. Our study uncovers that answer-related tokens, even while still masked, tend to follow more stable denoising trajectories (\ie less fluctuating confidence) when they eventually converge to correct answers. Then, (ii) examining whether the relatively less confident models in the early decoding phase can recover toward correct responses after receiving promising partially denoised sequences from more confident models, thereby allowing them to re-enter subsequent knowledge exchange.

Building upon our findings, we propose \textbf{TIE} (\textbf{T}rajectory-based \textbf{I}terative \textbf{E}nsembling), a knowledge fusion framework in which trajectories from more confident models are iteratively relayed to other models so that complementary knowledge from different models can be naturally integrated. 
Each model monitors the confidence dynamics of answer-related tokens, allowing the framework to identify which model is currently following a more reliable trajectory toward the correct response. Models whose confidence trajectories become unstable are provided with partially denoised sequences from more reliable counterparts and continue generation from those intermediate states.

Through this process, models that deviate from the correct trajectory can be guided back onto promising generation paths when provided with sufficiently reliable intermediate responses, as observed in our analysis. This knowledge transfer process is repeated periodically throughout generation. Interestingly, the model producing the more reliable response frequently changes across denoising steps, suggesting that different models contribute distinct strengths at different stages of generation. Consequently, all participating models collaboratively contribute to refining the final response.

Extensive experiments across diverse domains, including general reasoning, mathematics, coding, and planning, demonstrate that TIE consistently improves over individual MDLMs, highlighting the effectiveness of continual knowledge transfer guided by confidence dynamics over answer-related tokens. Our in-depth analyses further reveal that TIE is most effective when constituent models exhibit both comparable and strong individual capabilities. Overall, our findings provide practical guidelines for effective MDLM ensembling.

\section{Preliminaries}\label{sec:prelim}
\paragraph{Masked diffusion language models.}
Assume that $\mathbf{x}$ is a clean sequence consisting of a single token, $\mathbf{m}$ is the one-hot representation of the mask index, and $z_t$ denotes the token state at an intermediate noise level $t$. The forward process in Masked Diffusion Language Models (MDLMs) is defined as follows~\citep{austin2021structured}:
\begin{equation}~\label{eq:mdlm_forward_proces}
q(\mathbf{z}_t | \mathbf{x}) = \mathrm{Cat}(\mathbf{z}_t; \alpha_t\mathbf{x} + (1-\alpha_t)\mathbf{m}),
\end{equation}
where $\alpha_t$ is a predefined noise schedule. For an earlier level $s<t$, posterior distribution $q(\mathbf{z}_s | \mathbf{z}_t)$ can be analytically expressed. If $\mathbf{z}_t \neq \mathbf{m}$, the posterior is deterministic and satisfies
\begin{equation}
q(\mathbf{z}_s \mid \mathbf{z}_t,\mathbf{x})
=
\mathrm{Cat}(\mathbf{z}_s;\mathbf{z}_t).
\end{equation}
Otherwise, when $\mathbf{z}_t=\mathbf{m}$, the posterior becomes
\begin{equation}
\resizebox{0.95\columnwidth}{!}{$
q(\mathbf{z}_s \mid \mathbf{z}_t,\mathbf{x})
=
\mathrm{Cat}\!\left(
\mathbf{z}_s;
\dfrac{
(1-\alpha_s)\mathbf{m}
\!+\!
(\alpha_s\!-\!\alpha_t)\mathbf{x}
}{
1\!-\!\alpha_t
}
\right).
$}
\end{equation}
Following prior MDLM formulations~\citep{austin2021structured,shi2024simplified}, the model learns the reverse process by approximating the posterior distribution only on masked positions while unmasked tokens remain unchanged. Accordingly, the posterior transition is parameterized by a neural network $f_\theta$:
$
p_\theta(\mathbf{z}_s | \mathbf{z}_t)
:=
q(\mathbf{z}_s | \mathbf{z}_t,f_\theta(\mathbf{z}_t,t)),
$
where $f_\theta$ estimates the clean-token distribution conditioned on the noisy state $\mathbf{z}_t$ and diffusion time $t$. The training objective for a sequence of length $L$ is formulated as the negative evidence lower bound (ELBO): 
\begin{equation}
\vspace{-2mm}
\begin{aligned}
\mathcal{L}_\infty^{\text{ELBO}}
&=\int_0^1
\frac{\partial_t \alpha_t}{1-\alpha_t}
\,\mathbb{E}_{\mathbf{x}\sim q_0,\,\mathbf{z}_t\sim q_t(\mathbf{z}_t|\mathbf{x})} \\
&\qquad
\Bigg[\sum^{L}_{l:\mathbf{z}_t^{(l)}=\mathbf{m}}\mathbf{x}^{(l)}\cdot\log f_\theta^{(l)}(\mathbf{z}_t,t)\Bigg]dt,
\end{aligned}
\end{equation}
where $q_0$ represents the data distribution, and the summation is taken over masked positions.

\paragraph{Ancestral sampling and unmasking.}
During the inference phase, the diffusion process is discretized into $T$ denoising steps, with the sequence initialized as fully masked.
The model then iteratively denoises the sequence by sampling from the reverse process: $\mathbf{x}_{t-1}\sim p_{\theta}(\mathbf{x}_{t-1}|\mathbf{x}_{t})$ for $t= T, \ldots, 1$. Various policies for determining which tokens to unmask have been introduced in prior work, including confidence-based~\citep{kim2025train}, thresholding~\citep{wu2026fastdllm}, and KL-divergence criteria~\citep{kim2026klass}.

\begin{figure*}[t]
    \centering
    \includegraphics[width=\textwidth]{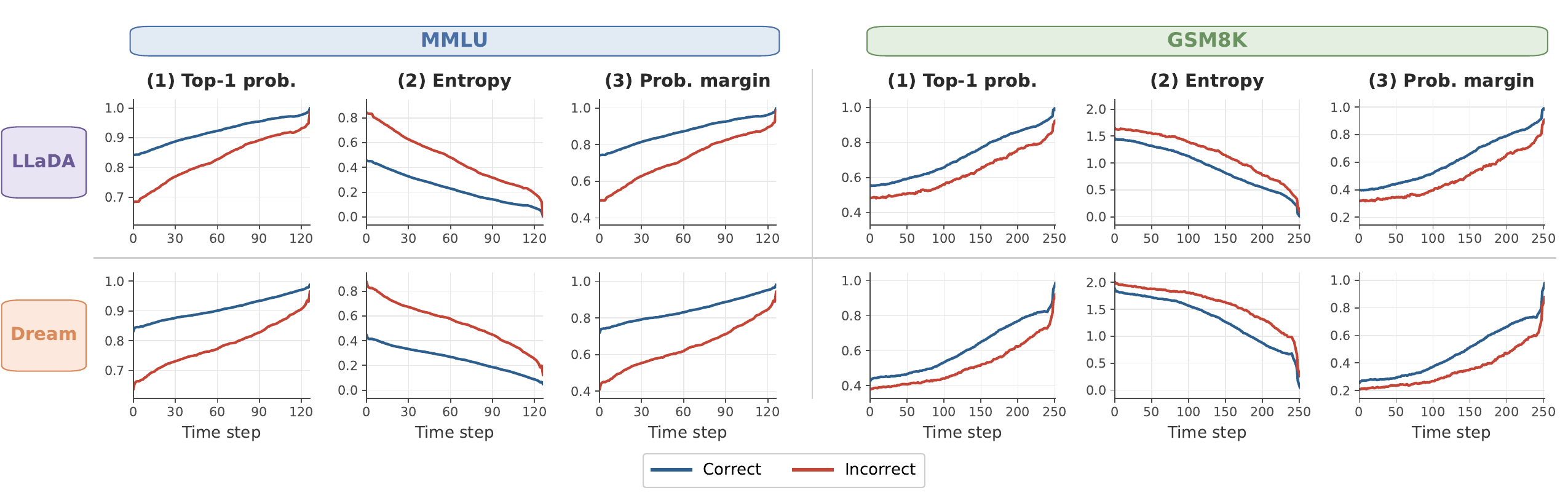}
    \vspace{-4mm}
    \caption{Evolution of three confidence metrics on answer tokens across decoding steps, averaged over answer positions and grouped by correct and incorrect samples. Correct samples consistently show higher confidence across all three metrics (higher top-1 confidence/confidence margin, lower entropy).}
    \label{fig:confidence}
    \vspace{-5mm}
\end{figure*}

\section{Towards Effective MDLM Ensembling}
\label{sec:observation}
We begin by presenting two observations on MDLM decoding dynamics that provide the key insights underlying our ensembling framework. Motivated by the goal of MDLM ensembling, we specifically focus on \emph{what characterizes high-quality MDLM decoding trajectories} by contrasting correct and incorrect samples. Our analysis reveals two key insights: (i) correct decoding trajectories show more stable and confident answer-token dynamics, and (ii) sharing even a partial portion of such trajectories can steer weaker models toward correct answers.

\paragraph{Analysis setup.}~The analysis is conducted on MMLU~\citep{mmlu} and GSM8K~\citep{gsm8k} using LLaDA-1.5~\citep{llada1.5} and Dream-7B-Instruct~\citep{dream}. We employ semi-autoregressive generation with a block size of 16 and low-confidence remasking. The generation length is set to 128 tokens for MMLU and 256 tokens for GSM8K.

\begin{table}[t]
\centering
\caption{Comparison of token change counts $\mathcal{C}^{(T)}$ between correct and incorrect samples.}
\vspace{-2mm}
\label{tab:answer_token_changes}
\resizebox{\columnwidth}{!}{%
\begin{tabular}{lcccc}
\toprule
\multirow{2}{*}{Models} & \multicolumn{2}{c}{MMLU} & \multicolumn{2}{c}{GSM8K} \\
\cmidrule(lr){2-3}\cmidrule(lr){4-5}
 & Correct & Incorrect & Correct & Incorrect \\
\midrule
LLaDA & 1.81 & 4.27 & 32.29 & 51.48 \\
Dream & 2.32 & 6.19 & 40.88 & 58.07 \\
\bottomrule
\end{tabular}
}
\vspace{-3mm}
\end{table}

\subsection{Correct Samples Are More Stable and Confident in Their Answers}
\label{subsec:observation1}
Our first observation investigates how the decoding dynamics of answer tokens differ between correct and incorrect samples. Our key question is: \textit{Do correctly answered samples show greater consistency and confidence in their answers during decoding?} 

To analyze this, we follow the generation setting of \citet{li2026diffusion}, which divides tokens into reasoning and answer tokens, where answer tokens are defined as the tokens appearing after the ``Answer:'' suffix. We then quantify the stability of answer tokens via the \emph{Token Change Count} ($\mathcal{C}^{(n)}$), defined as the total number of top-1 (\ie highest-probability) token changes between consecutive decoding steps, accumulated over $n$ decoding steps and the masked answer-token positions at each step $t$: 
\begin{equation}
\label{eq:tcc}
\resizebox{0.95\columnwidth}{!}{$\displaystyle
\mathcal{C}^{(n)}
= \sum_{t=T-n+1}^{T-1}
  \sum_{a \in \mathcal{A}^{(t)}}
  \mathbf{1}\!\left[
    \arg\max \mathbf{p}_a^{(t)} \neq \arg\max \mathbf{p}_a^{(t+1)}
  \right],
$}
\end{equation}
where $\mathcal{A}^{(t)}$ denotes the set of masked answer-token positions at step $t$, $T$ denotes the total number of decoding steps, and $\mathbf{p}_a^{(t)}$ denotes the predicted probability distribution at position $a$ and step $t$. A lower $\mathcal{C}^{(n)}$ indicates that answer tokens change less frequently throughout decoding, suggesting a more stable decoding behavior.

\Cref{tab:answer_token_changes} shows that $\mathcal{C}^{(T)}$ of incorrect samples is roughly twice that of correct samples, indicating that incorrect samples exhibit substantially less stability in their answers. For a detailed analysis, we examine how the confidence of answer tokens evolves during decoding using three metrics: (1) top-1 confidence, (2) entropy, and (3) confidence margin (the gap between the top-1 and top-2 probabilities). As shown in \Cref{fig:confidence}, correct samples consistently exhibit higher confidence on answer tokens throughout decoding across all three metrics. These results imply that confidence-based signals over answer tokens may serve as reliable indicators for identifying promising decoding trajectories.

\begin{table}[t]
\centering
\caption{Correction rate (\%) under different injection ratios. Each model continues decoding from a partial decoding trajectory generated by another model.}
\vspace{-2mm}
\label{tab:correction_rate}
\resizebox{\columnwidth}{!}{%
\begin{tabular}{llcc}
\toprule
Dataset & Model & Injection ratio & Correction rate \\
\midrule
\multirow{4}{*}{MMLU}
& LLaDA & \multirow{2}{*}{33\%} & 56.43 \\
& Dream &  & 68.84 \\
\cmidrule{2-4}
& LLaDA & \multirow{2}{*}{50\%} & 65.22 \\
& Dream &  & 76.57 \\
\midrule
\multirow{4}{*}{GSM8K}
& LLaDA & \multirow{2}{*}{33\%} & 74.66 \\
& Dream &  & 78.63 \\
\cmidrule{2-4}
& LLaDA & \multirow{2}{*}{50\%} & 72.60 \\
& Dream &  & 80.92 \\
\bottomrule
\end{tabular}
}
\vspace{-3mm}
\end{table}

\begin{figure*}[t]
\centering
\includegraphics[width=\textwidth]{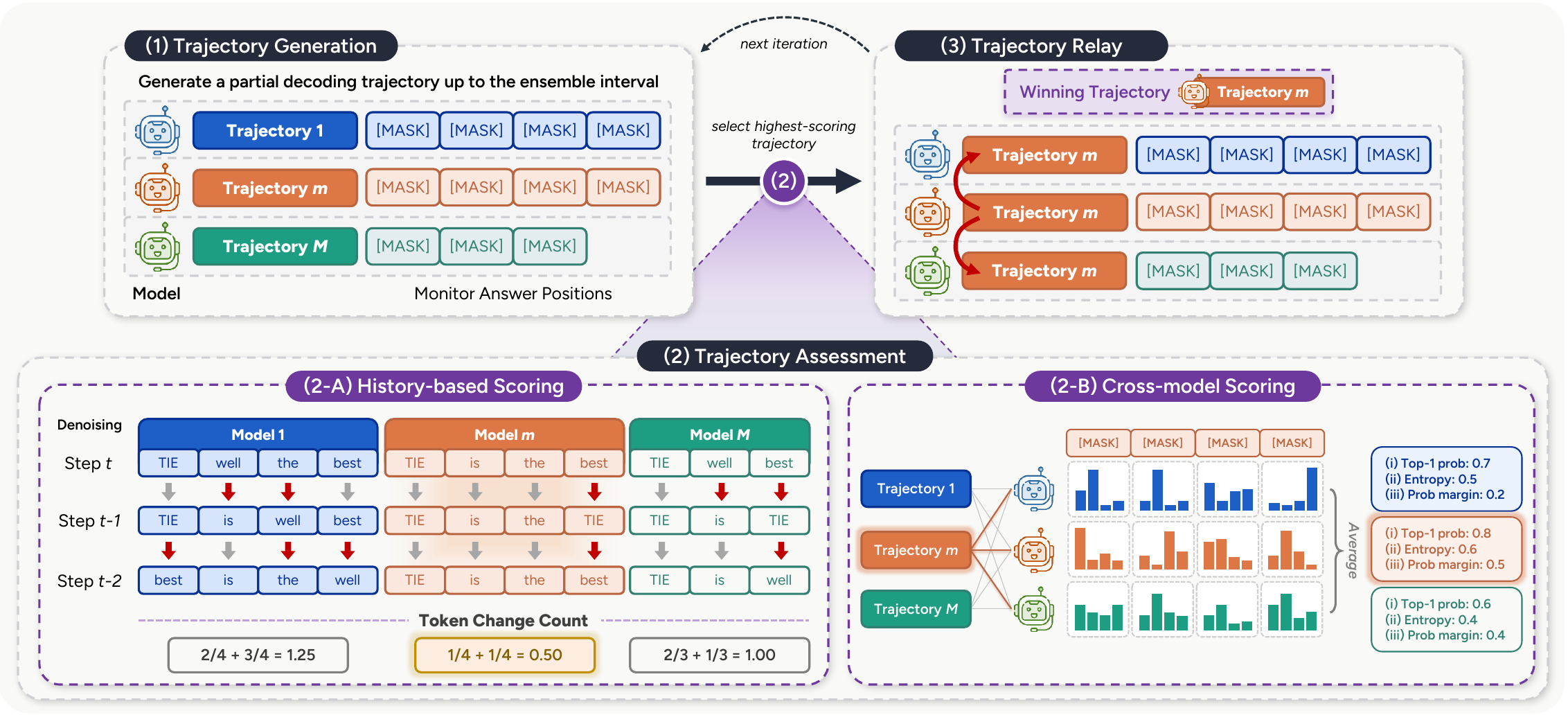}
\vspace{-3mm}
\caption{\textbf{Overview of TIE.} Each MDLM first independently generates a partial decoding trajectory from its current state. TIE then evaluates these trajectories using confidence-based scoring over answer-token positions, relays the most reliable trajectory across models, and continues decoding from the resulting more reliable intermediate state. In (2-A), the answer-token positions are originally masked, but we display their top-1 tokens for visualization.}
\vspace{-5mm}
\label{fig:main_figure}
\end{figure*}

\subsection{Reliable Partial Decoding Trajectories Guide Models Toward Correct Answers}
\label{subsec:observation2}
Given that promising decoding trajectories can be identified at early stages (\Cref{subsec:observation1}), we next examine whether such reliable decoding trajectories from one model can guide subsequent decoding of another toward a correct answer, which could be key to cross-model collaboration during decoding. 
Specifically, we investigate how often a model that initially generates an incorrect answer can be corrected when given an early portion of a decoding trajectory from another model that produced the correct answer. As shown in \Cref{tab:correction_rate}, we observe that a substantial portion of incorrect samples can be corrected even when only one-third of the correct decoding trajectory is provided. These results demonstrate that sharing reliable decoding trajectories among models can rectify generations that would otherwise produce incorrect answers. This motivates us to develop a trajectory-based ensemble framework in which models iteratively correct one another, enabling weaker models to recover rather than be discarded. As a result, different models can contribute their distinct strengths at different stages of generation.

\section{TIE: Trajectory-based Iterative Ensembling}\label{sec:method}
Guided by our two observations, we propose \textbf{TIE} (\textbf{T}rajectory-based \textbf{I}terative \textbf{E}nsembling), an MDLM-specific ensemble method that combines complementary strengths guided by confidence dynamics at answer-relevant positions. Given $M$ constituent MDLMs, TIE proceeds in a recurring three-step cycle: (i) \textbf{Trajectory Generation} (\Cref{subsec:method_1}) - each model independently performs unmasking for $n$ decoding steps;
(ii) \textbf{Trajectory Assessment} (\Cref{subsec:method_2}) - each trajectory is scored using confidence-based metrics;
(iii) \textbf{Trajectory Relay} (\Cref{subsec:method_3}) - the highest-scoring trajectory is relayed to all constituent models, replacing their current decoding states to recover models from erroneous or suboptimal trajectories for more effective collaboration in subsequent ensemble steps. This cycle repeats until generation terminates, progressively steering the ensemble toward a more accurate final response.

\paragraph{Notation.}
Throughout this section, subscripts $m$ index constituent models and superscripts $(t)$ index decoding steps. For example, $\mathcal{T}_m^{(n)}$ denotes the trajectory $\mathcal{T}$ of model $m$ after $n$ steps.

\subsection{Trajectory Generation}
\label{subsec:method_1}
In each round, every constituent model $m \in [M]$ independently decodes from its current state for $n$ steps, where $n$ denotes the \emph{ensemble interval}, producing a partial decoding trajectory $\mathcal{T}_m^{(n)}$. During decoding, answer-token positions are unmasked only after all reasoning-token positions have been fully unmasked. The effect of ensemble interval is studied in \Cref{subsec:ablation}.

\subsection{Trajectory Assessment}
\label{subsec:method_2}
Given the $M$ partial trajectories $\{\mathcal{T}_m^{(n)}\}_{m=1}^{M}$ from each model, this step assigns a confidence-based score to each trajectory in order to identify the most reliable one. Following \Cref{subsec:observation1}, we consider four scoring metrics:
(i) negative token change count,
(ii) top-1 confidence,
(iii) negative entropy, and
(iv) confidence margin, where higher values indicate greater confidence. All metrics are computed only over masked answer-token positions $\mathcal{A}_m^{(t)}$ to identify the trajectory with the most stable and confident decoding behavior over answer tokens.
Each scoring function is applied independently, and their effectiveness is compared in \Cref{subsec:ablation}.

\noindent\textbf{History-based scoring for the token change count.}
The token change count $\mathcal{C}_m^{(n)}$ reflects the history of answer-token stability throughout the generation of $\mathcal{T}_m^{(n)}$. To compute $\mathcal{C}_m^{(n)}$, we track how many masked answer-token positions $\mathcal{A}_m^{(t)}$ change their top-1 tokens across decoding steps during trajectory generation (see \Cref{eq:tcc}). Since $|\mathcal{A}_m^{(t)}|$ may differ across models and decoding steps, we normalize $\mathcal{C}_m^{(n)}$ by $|\mathcal{A}_m^{(t)}|$ to prevent the token change count from inflating simply due to having more answer tokens:
\begin{equation}
\label{eq:tcc_score}
\mathrm{Score}\bigl(\mathcal{T}_m^{(n)}\bigr) \;=\; -\tilde{\mathcal{C}}_m^{(n)},
\end{equation}
where
\begin{equation}
\label{eq:tcc_score_norm}
\resizebox{0.95\columnwidth}{!}{$\displaystyle
\tilde{\mathcal{C}}_m^{(n)} \\
= \sum_{t=T-n+1}^{T-1} \frac{1}{|\mathcal{A}_m^{(t)}|}
   \sum_{a \in \mathcal{A}_m^{(t)}}
   \mathbf{1}\!\left[
     \arg\max \mathbf{p}_a^{(t)} \neq \arg\max \mathbf{p}_a^{(t+1)}
   \right].
$}
\end{equation}

\noindent\textbf{Cross-model scoring for logit-based metrics.}
Logit-based scoring functions (\ie top-1 confidence, entropy, and confidence margin) directly reflect the confidence over answer tokens for each $\mathcal{T}_m^{(n)}$. Since models can differ in their confidence calibration, we employ cross-model scoring. We forward each $\mathcal{T}_m^{(n)}$ through all constituent models and use the averaged score across models as its final confidence score:
\begin{equation}
\mathrm{Score}\bigl(\mathcal{T}_m^{(n)}\bigr) \;=\; \frac{1}{M} \sum_{m'=1}^{M} f\bigl(\mathcal{T}_m^{(n)};\, m'\bigr),
\label{eq:cross_model_score}
\end{equation}
where $f(\cdot;\, m')$ denotes the scoring function evaluated under the $m'$-th model. This procedure ensures the selection of a trajectory that is consistently confident across models rather than merely the one favored by its source model. We validate the effectiveness of cross-model scoring in Appendix~\ref{sec:appendix_cross_model}.

\subsection{Trajectory Relay}
\label{subsec:method_3}
The trajectory with the highest score is relayed to all models to recover those that previously produced suboptimal trajectories:
\begin{equation}
m^* \;=\; \arg\max_{m \in [M]} \mathrm{Score}\bigl(\mathcal{T}_m^{(n)}\bigr),
\label{eq:trajectory_selection}
\end{equation}
by replacing the current partial trajectory of each model with $\mathcal{T}_{m^*}^{(n)}$:
\begin{equation}
\mathcal{T}_m^{(n)} \leftarrow \mathcal{T}_{m^*}^{(n)} \quad \forall m \in [M].
\label{eq:distillation}
\end{equation}
After the relay, each model resumes independent decoding for the next $n$ steps. Since the prior decoding histories have been replaced, the trajectory scores are reset accordingly. This cycle of generation, assessment, and relay repeats until any of the constituent models completes decoding.

\subsection{Final Response Selection}
\label{subsec:method_4}
After decoding terminates, the constituent models produce $M$ candidate responses. We select the final answer as the response that exhibited the most stable answer-token dynamics throughout decoding, \ie the one with the lowest $\tilde{\mathcal{C}}_m^{(T)}$, breaking ties using the highest top-1 confidence. We ablate different selection strategies in Appendix~\ref{sec:appendix_response_selection}.

\begin{table*}[t]
\centering
\small
\setlength{\tabcolsep}{4pt}
\caption{Benchmark results for ensembling LLaDA and Dream. Post-generation ensembling with TIE (denoted as TIE-Post) corresponds to a special case of TIE where the ensemble interval equals the generation length. Dream is used as an aggregator for MoA. TIE uses token change count for trajectory assessment.}
\vspace{-0.1in}
\label{tab:main_results}
\resizebox{\textwidth}{!}{%
\begin{tabular}{l cccc cc c c}
\toprule
 & \multicolumn{4}{c}{\textbf{General}} & \multicolumn{2}{c}{\textbf{Math}} & \textbf{Coding} & \textbf{Planning} \\
\cmidrule(lr){2-5}\cmidrule(lr){6-7}\cmidrule(lr){8-8}\cmidrule(lr){9-9}
\textbf{Method} & MMLU & \makecell{MMLU$^*$\\(high-perf.)} & ARC-C & WinoGrande & GSM8K & MATH500 & MBPP & Countdown \\
\cmidrule[1.5pt]{1-9}
\multicolumn{9}{l}{\textit{Single Models}} \\
\cmidrule[1.5pt]{1-9}
LLaDA & 61.23 & 71.45 & 85.15 & 71.59 & 78.77 & 37.4 & 53.16 & 13.4 \\
Dream    & 67.46 & 77.37 & 86.69 & 72.22 & 78.39 & 48.0 & 63.23 & 16.4 \\
\cmidrule[1.5pt]{1-9}
\multicolumn{9}{l}{\textit{Post-generation Ensemble}} \\
\cmidrule[1.5pt]{1-9}
PairRM~\citep{jiang-etal-2023-llm} & 65.90 & 75.96 & 86.95 & 73.24 & 81.43 & 43.2 & 55.97 & 18.0 \\
MoA~\citep{wang2025mixtureofagents} & 65.08 & 74.89 & 84.39 & 71.67 & 74.53 & 39.6 & 61.83 & 16.0 \\
\oursrow TIE-Post \textbf{(Ours)}
& 67.26 & 77.92 & 88.57 & \textbf{73.72} & 80.29 & 43.0 & 62.53 & \textbf{19.2} \\
\cmidrule[1.5pt]{1-9}
\multicolumn{9}{l}{\textit{Intermediate-generation Ensemble}} \\
\cmidrule[1.5pt]{1-9}
\oursrow TIE \textbf{(Ours)}
& \textbf{67.55} & \textbf{78.12} & \textbf{89.16} & 72.85 & \textbf{83.47} & \textbf{48.6} & \textbf{64.17} & 18.8 \\
\bottomrule
\end{tabular}%
}
\vspace{-0.12in}
\end{table*}

\begin{table}[t]
\centering
\small
\setlength{\tabcolsep}{5pt}
\caption{Results for ensembling DreamCoder and DiffuCoder on MBPP. DreamCoder is used as an aggregator for MoA.}
\vspace{-0.1in}
\label{tab:dream_diffucoder}
\begin{tabularx}{\columnwidth}{X c}
\toprule
 \textbf{Method} & MBPP \\
\cmidrule[1.5pt]{1-2}
\multicolumn{2}{l}{\textit{Single Models}} \\
\cmidrule[1.5pt]{1-2}
DreamCoder-7B-Instruct & 75.88  \\
DiffuCoder-7B-Instruct  & 72.60\\
\cmidrule[1.5pt]{1-2}
\multicolumn{2}{l}{\textit{Post-generation Ensemble}} \\
\cmidrule[1.5pt]{1-2}
PairRM~\citep{jiang-etal-2023-llm} & 76.11 \\
MoA~\citep{wang2025mixtureofagents} & 76.58 \\
\cmidrule[1.5pt]{1-2}
\multicolumn{2}{l}{\textit{Intermediate-generation Ensemble}} \\
\cmidrule[1.5pt]{1-2}
\oursrow TIE \textbf{(Ours)} & \textbf{77.52} \\
\bottomrule
\end{tabularx}
\vspace{-0.1in}
\end{table}

\section{Experiments}
\label{sec:experiments}

In this section, we demonstrate the effectiveness of TIE through extensive experiments. We first describe our experimental setup (\Cref{subsec:exper_settings}), then present three key analyses (\Cref{subsec:main_analysis}), and conclude with an ablation study (\Cref{subsec:ablation}).

\subsection{Experimental Settings}\label{subsec:exper_settings}

\noindent\textbf{Models.} We evaluate our method using four widely-used MDLMs: LLaDA-1.5~\citep{llada1.5}, Dream-7B-Instruct~\citep{dream}, DreamCoder-7B-Instruct~\citep{dreamcoder}, and DiffuCoder-7B-Instruct~\citep{gong2026diffucoder}.

\noindent\textbf{Benchmarks.} To assess generalization across diverse domains, we evaluate TIE on seven benchmarks spanning four task categories: (i) \emph{general reasoning}: MMLU~\citep{mmlu}, ARC-Challenge~\citep{arc}, and WinoGrande~\citep{winogrande}; (ii) \emph{mathematical reasoning}: GSM8K~\citep{gsm8k} and MATH500~\citep{math500}; (iii) \emph{coding}: MBPP-sanitized~\citep{mbpp}; and (iv) \emph{planning}: Countdown\footnote{\url{https://huggingface.co/datasets/predibase/countdown}}.

\noindent\textbf{Baselines.} 
We compare our method against two representative post-generation ensemble methods originally proposed for autoregressive LLMs: PairRM~\citep{jiang-etal-2023-llm}, which ranks candidate responses, and MoA~\citep{wang2025mixtureofagents}, which synthesizes candidate responses with an aggregator model. Both are applied to the responses independently generated by each MDLM. We further categorize our method according to when ensembling occurs: \emph{post-generation ensemble} and \emph{intermediate-generation ensemble}. Post-generation ensemble selects the final answer from independently generated responses following \Cref{subsec:method_4}. This corresponds to a special case of TIE where the ensemble interval equals the total generation length. Intermediate-generation ensemble, in contrast, enables models to collaborate during decoding. Unless otherwise stated, TIE uses token change count for trajectory assessment; we compare it with the other scoring functions (top-1 confidence, entropy, and confidence margin) in \Cref{subsec:ablation}.

\noindent\textbf{Implementation.} We adopt semi-autoregressive generation with a block size of 16 tokens, low-confidence remasking, and greedy decoding. The generation length is set to 128 tokens for general-reasoning tasks, 256 tokens for GSM8K and Countdown, and 512 tokens for MATH500 and coding benchmarks. Except for coding tasks, all models are prompted to provide reasoning before the final answer~\citep{cot}. The ensemble interval is set to 16 decoding steps by default, and set to 32 steps when using token change count on general reasoning, coding, and planning. During ensembling, all constituent models perform generation in parallel, resulting in generation latency comparable to that of a single model.

\begin{table*}[t]
\centering
\small
\setlength{\tabcolsep}{4pt}
\caption{Comparison of scoring criteria for trajectory assessment when ensembling LLaDA and Dream with TIE.}
\vspace{-0.1in}
\label{tab:ab_scoring}
\resizebox{\textwidth}{!}{%
\begin{tabular}{l cccc cc c c}
\toprule
 & \multicolumn{4}{c}{\textbf{General}} & \multicolumn{2}{c}{\textbf{Math}} & \textbf{Coding} & \textbf{Planning} \\
\cmidrule(lr){2-5}\cmidrule(lr){6-7}\cmidrule(lr){8-8}\cmidrule(lr){9-9}
\textbf{Scoring criteria} & MMLU & \makecell{MMLU$^*$\\(high-perf.)} & ARC-C & WinoGrande & GSM8K & MATH500 & MBPP & Countdown \\
\midrule
Token change count & \textbf{67.55} & \textbf{78.12} & \textbf{89.16} & 72.85 & 83.47 & \textbf{48.6} & \textbf{64.17} & 18.8 \\
Top-1 confidence & 67.25 & 77.69 & 88.82 & \textbf{73.95} & 82.71 & 47.0 & 62.06 & 18.8 \\
Entropy & 67.53 & \textbf{78.12} & 88.57 & 71.90 & \textbf{83.62} & 45.2 & 62.76 & \textbf{19.4} \\
Confidence margin & 67.34 & 77.61 & 88.74 & 73.88 & 82.56 & 48.4 & 62.06 & 18.6 \\
\bottomrule
\end{tabular}%
}
\vspace{-0.15in}
\end{table*}

\subsection{Main Analysis}\label{subsec:main_analysis}
\paragraph{(i) TIE selects better decoding trajectories across domains.}
\Cref{tab:main_results} shows that TIE improves over individual models across a wide range of domains, demonstrating that confidence-based scoring over answer tokens can effectively identify high-quality decoding trajectories. TIE also outperforms existing post-generation ensemble methods such as PairRM and MoA on most benchmarks, without requiring an additional reward model or aggregation stage. The benefits of TIE extend beyond LLaDA and Dream: as shown in \Cref{tab:dream_diffucoder}, TIE also outperforms individual models when applied to code-specialized MDLMs such as DreamCoder and DiffuCoder.

\paragraph{(ii) TIE is most effective with comparable and strong constituent models.}
Although TIE generally outperforms individual models, the gains are not uniform across all benchmarks. In particular, when one model substantially underperforms the others, the ensemble becomes less effective, as weaker models may introduce noisy signals during trajectory aggregation. For example, on MATH500, where LLaDA and Dream exhibit a performance gap above 10\%, TIE yields only a marginal gain over Dream. 
We also observe that the gains from TIE increase as the performance of the constituent models improves. TIE achieves especially large improvements on ARC-C and GSM8K, where the individual models already perform well. To further examine this, we define MMLU$^*$ as the subset of MMLU subjects on which both LLaDA and Dream achieve over 60\% accuracy. On this subset, TIE yields larger improvements than on the full MMLU. We attribute this to the more reliable confidence dynamics of stronger models - they are more accurate in knowing what they know and what they do not - which TIE directly leverages for trajectory assessment. These results suggest that ensembling becomes more effective when constituent models have comparable and strong individual capabilities.

\paragraph{(iii) TIE allows different models to contribute at different decoding stages.} One interesting finding is that the model producing the highest-scoring trajectory changes dynamically throughout the ensembling process, rather than a single model consistently leading the generation: with token change count, the leading model changes at 10.55--32.04\% of trajectory relay steps across benchmarks (\Cref{tab:model_change_rate} in Appendix~\ref{sec:appendix_leading_change}). Consistent with \Cref{subsec:observation2}, this confirms that models initially heading toward incorrect answers can recover after receiving reliable trajectories from other models and contribute high-quality trajectories in later aggregation steps. 
Consequently, repeated trajectory aggregation during decoding allows TIE to leverage the strengths of different models at different stages, progressively converging toward better final answers. This explains why intermediate-generation ensembling outperforms post-generation ensembling in \Cref{tab:main_results}.

\begin{table}[t]
\centering
\small
\setlength{\tabcolsep}{3pt}
\caption{Ablation on the ensemble interval $n$. Trajectory assessment is based on token change count.}
\vspace{-0.1in}
\label{tab:steps_ablation}
\resizebox{\columnwidth}{!}{%
\begin{tabular}{l ccc cc}
\toprule
 & \multicolumn{3}{c}{\textbf{General}} & \multicolumn{2}{c}{\textbf{Math}} \\
\cmidrule(lr){2-4}\cmidrule(lr){5-6}
\textbf{Method} & MMLU$^*$ & ARC-C & Wino. & GSM8K & MATH500 \\
\midrule
LLaDA & 71.45 & 85.15 & 71.59 & 78.77 & 37.4 \\
Dream & 77.37 & 86.69 & 72.22 & 78.39 & 48.0 \\
\midrule
\oursrow TIE ($n{=}8$)  & \textbf{78.21} & 89.33 & 71.90 & 82.34 & 46.0 \\
\oursrow TIE ($n{=}16$) & 78.17 & \textbf{89.51} & \textbf{73.32} & \textbf{83.47} & \textbf{48.6} \\
\oursrow TIE ($n{=}32$) & 78.12 & 89.16 & 72.85 & 82.49 & 46.0 \\
\bottomrule
\end{tabular}%
}
\vspace{-0.1in}
\end{table}

\begin{table*}[t]
\centering
\small
\setlength{\tabcolsep}{4pt}
\caption{Benchmark results under different inference acceleration strategies for MDLMs. (a) \textit{Threshold} unmasks all tokens with confidence above a fixed threshold $\tau$. (b) \textit{Top-$k$} unmasks the $k$ most confident tokens per step.}
\label{tab:mdlm-acceleration}

\begin{subtable}{0.48\textwidth}
\centering
\caption{Threshold ($\tau{=}0.9$)}
\label{tab:mdlm-threshold}
\resizebox{\linewidth}{!}{%
\begin{tabular}{l c c c c}
\toprule
 & \multicolumn{2}{c}{\textbf{General}} & \textbf{Math} & \textbf{Coding} \\
\cmidrule(lr){2-3}\cmidrule(lr){4-4}\cmidrule(lr){5-5}
\textbf{Method} & MMLU$^*$ & ARC-C & GSM8K & MBPP \\
\midrule
\multicolumn{5}{l}{\textit{Single Models}} \\
\midrule
LLaDA & 71.34 & 84.47 & 80.44 & 53.63 \\
Dream    & 77.14 & 86.69 & 78.39 & \textbf{63.00} \\
\midrule
\multicolumn{5}{l}{\textit{Intermediate-generation Ensemble}} \\
\midrule
\oursrow TIE (Token change count)
            & 77.56 & \textbf{89.33} & 82.26 & \textbf{63.00} \\
\oursrow TIE (Top-1 confidence)
            & \textbf{77.75} & 88.23 & \textbf{83.24} & 62.76 \\
\bottomrule
\end{tabular}%
}
\end{subtable}
\hfill
\begin{subtable}{0.48\textwidth}
\centering
\caption{Top-$k$ ($k{=}2$)}
\label{tab:mdlm-topk}
\resizebox{\linewidth}{!}{%
\begin{tabular}{l c c c c}
\toprule
 & \multicolumn{2}{c}{\textbf{General}} & \textbf{Math} & \textbf{Coding} \\
\cmidrule(lr){2-3}\cmidrule(lr){4-4}\cmidrule(lr){5-5}
\textbf{Method} & MMLU$^*$ & ARC-C & GSM8K & MBPP \\
\midrule
\multicolumn{5}{l}{\textit{Single Models}} \\
\midrule
LLaDA & 71.25 & 84.81 & 78.92 & 46.37 \\
Dream   & 76.11 & 86.26 & 73.77 & 53.86 \\
\midrule
\multicolumn{5}{l}{\textit{Intermediate-generation Ensemble}} \\
\midrule
\oursrow TIE (Token change count)
            & 77.32 & 89.16 & 80.14 & \textbf{58.31} \\
\oursrow TIE (Top-1 confidence)
            & \textbf{77.88} & \textbf{89.93} & \textbf{81.58} & 55.27 \\
\bottomrule
\end{tabular}%
}
\end{subtable}
\vspace{-0.2in}
\end{table*}

\subsection{Ablation Study}\label{subsec:ablation}

We provide three ablations in this section: (i) the choice of scoring function for trajectory assessment, (ii) the effect of varying the ensemble interval $n$, and (iii) the compatibility of TIE with existing MDLM decoding acceleration methods. Additional ablations are provided in Appendices~\ref{sec:appendix_tcc_norm} to \ref{sec:appendix_anchor_free}.

\paragraph{Comparison of scoring criteria.}
As shown in \Cref{tab:ab_scoring}, all scoring functions improve over individual models on most benchmarks. Token change count achieves the most robust performance, yielding the best results on five out of eight benchmarks. We attribute this to its ability to track the full decoding history, allowing it to better capture the stability of answer tokens throughout decoding.

\paragraph{Robustness across ensemble intervals.}
\Cref{tab:steps_ablation} shows the results under different ensemble intervals $n$. Across all settings, TIE outperforms individual models on most domains. Nevertheless, setting an appropriate ensemble interval is important. When the ensemble interval is too small, partially decoded trajectories may not contain sufficient information for reliable assessment. Conversely, too large intervals reduce the frequency of trajectory aggregation, limiting knowledge fusion across models. Overall, TIE remains robust within a reasonable range of 8 to 32 steps.

\paragraph{Compatibility with decoding acceleration.}
Since MDLMs are commonly combined with acceleration techniques for efficient inference, we further examine whether TIE remains effective when combined with such methods. We consider two standard acceleration techniques that serve as the basis of many existing MDLM inference acceleration methods: (i) \emph{thresholding}, which unmasks all tokens whose top-1 confidence exceeds a predefined threshold at each step, and (ii) \emph{top-$k$ unmasking} ($k>1$), which unmasks the top-$k$ confident tokens per step. As shown in \Cref{tab:mdlm-acceleration}, TIE remains consistently effective even when combined with these acceleration strategies, continuing to outperform individual models across multiple domains. These results demonstrate that TIE can be readily combined with existing acceleration techniques for more practical deployment.

\section{Related Work}
\label{sec:related work}

\paragraph{Autoregressive Language Models Ensemble.}
Owing to the prevailing success of autoregressive language models, prior work on LLM ensembling has been developed specifically for the autoregressive generation setting. These approaches can be broadly categorized according to the granularity at which model outputs are aggregated.

\paragraph{Output-level Ensemble.} These methods first elicit complete responses from each model independently and then aggregate them into a single answer. Early studies explored iterative multi-agent debate~\citep{du2024improving,chen-etal-2024-reconcile}, whereas more recent approaches instead directly fuse independently generated responses. 
For example, LLM-Blender~\citep{jiang-etal-2023-llm} trains a dedicated fuser to synthesize a final answer, while ~\citet{si-etal-2023-getting} trains a classifier to select the optimal response. MoA~\citep{wang2025mixtureofagents} designates one constituent model as an aggregator that consolidates the outputs of the others. Although effective, these methods incur additional inference costs for the fusion stage, cannot integrate models' knowledge during the generation process itself, and typically rely on a large pool of candidate responses.

\paragraph{Span-/Token-level Ensemble.} Unlike output sequence-level methods, these approaches perform aggregation during generation at finer granularities. Span-level methods~\citep{liu-etal-2025-cool,xu-etal-2025-hit} iteratively construct the final response by selecting the most promising span (\eg, a sequence of words or tokens) among candidate spans proposed by multiple models, often based on perplexity from other models. Token-level methods~\citep{yu-etal-2024-breaking,xu-etal-2024-bridging,yao2025determinethenensemble,yun2026when} further refine the aggregation granularity to individual tokens, aggregating next-token probability distributions from multiple models and sampling from the aggregated distribution.

Both span-level and token-level ensemble methods require the next span or token to be at the same position across participating models in order to aggregate them. Consequently, they are not directly applicable to settings in which responses are not generated autoregressively or in which the token generation order varies across models, where the notion of a ``next-token'' is not well-defined.

\section{Conclusion}
We introduced \textit{TIE}, a knowledge fusion framework that rethinks how heterogeneous Masked Diffusion Language Models can collaborate. Through continual intermediate exchange guided by confidence dynamics over answer-related tokens, \textit{TIE} allows models to recover from suboptimal trajectories and contribute complementary strengths throughout denoising. Without additional training, \textit{TIE} consistently improves performance across diverse reasoning tasks, highlighting the promise of collaborative inference for diffusion language models. We believe this work takes a meaningful step toward more effective MDLM orchestration.

\section*{Limitations}
Although TIE has been shown to be effective across various domains and generation settings, several aspects remain open for improvement. First, as discussed in \Cref{subsec:main_analysis}, TIE becomes less effective when the performance gap between constituent models is excessively large (\eg, greater than 10\%). This is a common challenge in LLM ensembling~\citep{yao2025determinethenensemble,yun2026when}, and our method could be further strengthened by incorporating mechanisms such as model routing, which selects suitable constituent models prior to ensembling. Second, our experiments mainly focus on ensembles of two models. Although Appendix~\ref{sec:appendix_three_models} shows that TIE further benefits from incorporating additional models, a more comprehensive investigation of how the effectiveness of MDLM ensembling scales with the number of constituent models remains a valuable direction for future work.

\bibliography{custom}

\clearpage
\appendix

\FloatBarrier
\section{Experimental Details}
\subsection{Dataset Details}

\paragraph{Selected MMLU subjects.}
In \Cref{subsec:main_analysis}, to examine whether ensembling stronger models yields larger gains, we define MMLU$^*$ as the subset of MMLU subjects on which both LLaDA and Dream achieve over 60\% accuracy. This subset comprises the following 35 subjects: astronomy, business ethics, clinical knowledge, college biology, college medicine, computer security, conceptual physics, elementary mathematics, high school biology, high school computer science, high school European history, high school geography, high school government and politics, high school macroeconomics, high school microeconomics, high school psychology, high school US history, high school world history, human aging, human sexuality, international law, jurisprudence, logical fallacies, management, marketing, medical genetics, miscellaneous, nutrition, philosophy, prehistory, public relations, security studies, sociology, US foreign policy, and world religions. 

\paragraph{Dataset splits.}
When ground-truth answers are available for the test split, we evaluate on the test split; otherwise, we use the validation split. For MBPP, we use the sanitized version, which filters out low-quality samples.

\paragraph{License.} All datasets and models used in the experiments, when accompanied by a license, permit their use for research purposes. Detailed information is provided in their respective references.

\subsection{Hardware} When ensembling models with TIE, each model is loaded onto a separate RTX 3090 GPU with bfloat16 precision.

\subsection{Prompts}
We present the prompt templates used in our experiments. For the multiple-choice and math domains, we adopt the prompt format from \texttt{simple-evals}\footnote{\url{https://github.com/openai/simple-evals}}. For Countdown, we follow the template used in \citet{wang2026spg}. We use the following prompt template for MBPP:

\begin{small}
\begin{verbatim}
{Question}

Your code should satisfy these tests:

{Tests}

Return only the Python function, no explanation.
\end{verbatim}
\end{small}

\subsection{Answer-token positions}
\label{sec:appendix_answer_token_pos}
We detail how answer-token positions are defined in our method. We append an \texttt{Answer:} suffix to the token sequence, and define the positions preceding the suffix as reasoning positions and those following it as answer positions. For coding tasks, we replace the \texttt{Answer:} suffix with \verb|```python|. During decoding, the answer positions are unmasked only after all reasoning positions have been fully unmasked.

To obtain more reliable confidence dynamics over answer positions, we exclude any answer position whose top-1 token is the end-of-sequence (EOS) token, as such positions are unrelated to the model's actual answer. Moreover, in domains such as code generation, answer sequences can become substantially longer, causing later answer tokens to become noisy and less accurate. We therefore evaluate confidence dynamics only on the first four masked answer positions in these domains. Consequently, the set of answer-token positions is updated dynamically at each decoding step for each model, as it may vary across both decoding steps and models.

\FloatBarrier
\section{When Does the Ensemble-Leading Model Change?}
\label{sec:appendix_leading_change}

\begin{figure}[!ht]
    \centering
    \begin{subfigure}[t]{0.49\columnwidth}
        \centering
        \includegraphics[width=\linewidth]{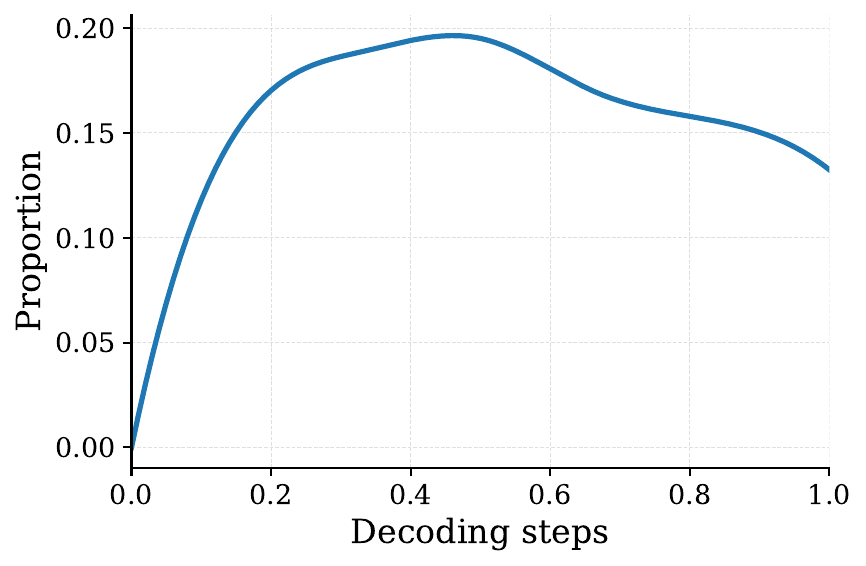}
        \caption{MMLU}
        \label{fig:dist_mmlu}
    \end{subfigure}
    \hfill
    \begin{subfigure}[t]{0.49\columnwidth}
        \centering
        \includegraphics[width=\linewidth]{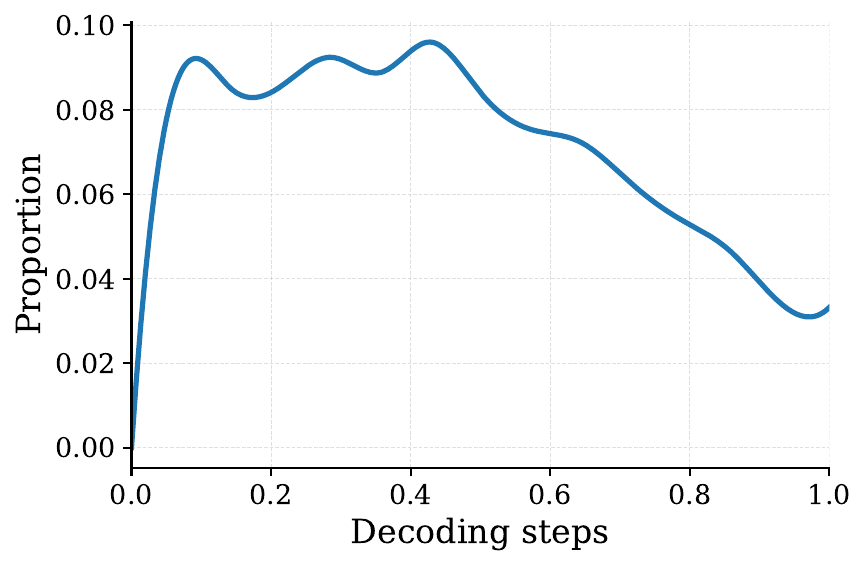}
        \caption{GSM8K}
        \label{fig:dist_gsm8k}
    \end{subfigure}

    \caption{Visualization of when the ensemble-leading model changes during decoding. The y-axis represents the proportion (\%) of decoding steps in which the highest-scoring model changes.}
    \label{fig:distributions}
\end{figure}

\begin{table*}[!t]
\centering
\small
\setlength{\tabcolsep}{5pt}
\caption{Model change rate (\%), defined as the percentage of trajectory-relay steps at which the selected (highest-scoring) model differs from the one selected at the previous step. A high change rate indicates that no single model dominates throughout decoding; instead, different models contribute at different stages.}
\label{tab:model_change_rate}
\vspace{-1mm}
\begin{tabular}{l ccc cc c}
\toprule
 &  \multicolumn{3}{c}{\textbf{General}} & \multicolumn{2}{c}{\textbf{Math}} & \textbf{Planning} \\
\cmidrule(lr){2-4}\cmidrule(lr){5-6} \cmidrule(lr){7-7}
 \textbf{Method} & MMLU & ARC-C & WinoGrande & GSM8K & MATH500 & Countdown \\
\midrule
Token change count & 12.28 & 10.55 & 12.59 & 23.74 & 27.80 & 32.04 \\
Top-1 confidence & 34.25 & 30.31 & 40.99 & 44.80 & 47.32 & 44.43 \\
\bottomrule
\end{tabular}
\vspace{-4mm}
\end{table*}

\Cref{tab:model_change_rate} shows that the model producing the highest-scoring trajectory changes frequently throughout generation. One natural question is at which stages of decoding these changes primarily occur. As shown in \Cref{fig:distributions}, we observe that the ensemble-leading model changes more frequently during the early decoding stages. This enables knowledge exchange among models to occur primarily during the early decoding steps, before a clear answer trajectory has fully emerged, allowing TIE to progressively converge toward higher-quality responses.

\section{Effect of Token Change Count Normalization}
\label{sec:appendix_tcc_norm}

\begin{table}[!htb]
\centering
\small
\setlength{\tabcolsep}{4pt}
\caption{Effect of normalizing the token change count by the number of masked answer-token positions.}
\label{tab:ab_token_change_count_norm}
\begin{tabular}{l cc}
\toprule
\textbf{Method} & GSM8K & MATH500 \\
\cmidrule[1.5pt]{1-3}
\multicolumn{3}{l}{\textit{Single Models}} \\
\cmidrule[1.5pt]{1-3}
LLaDA & 78.77 & 37.4   \\
Dream & 78.39 & 48.0  \\
\cmidrule[1.5pt]{1-3}
\multicolumn{3}{l}{\textit{Intermediate-generation Ensemble}} \\
\cmidrule[1.5pt]{1-3}
TIE w/o norm. & 82.64 & 47.8 \\
TIE & \textbf{83.47} & \textbf{48.6} \\
\bottomrule
\end{tabular}%
\end{table}

In the history-based trajectory assessment (\Cref{subsec:method_2}), we use the normalized token change count $\tilde{\mathcal{C}}_m^{(n)}$, which normalizes $\mathcal{C}_m^{(n)}$ by the number of masked answer-token positions $|\mathcal{A}_m^{(t)}|$, instead of directly using $\mathcal{C}_m^{(n)}$. This normalization is important because $|\mathcal{A}_m^{(t)}|$ can differ across models, and using $\mathcal{C}_m^{(n)}$ without accounting for this difference may fail to reflect answer-token stability accurately. In particular, a model with a larger $|\mathcal{A}_m^{(t)}|$ may naturally exhibit a higher $\mathcal{C}_m^{(n)}$ simply due to having more answer-token positions, rather than due to genuinely unstable decoding dynamics. \Cref{tab:ab_token_change_count_norm} shows the effect of normalizing $\mathcal{C}_m^{(n)}$. As shown in the table, compensating for differences in $|\mathcal{A}_m^{(t)}|$ leads to improved performance.

\section{Effect of Cross-Model Scoring}
\label{sec:appendix_cross_model}

\begin{table*}[!t]
\centering
\small
\setlength{\tabcolsep}{4pt}
\caption{Comparison between TIE with and without cross-model scoring. \textit{TIE w/o cross-model scoring} (\ie source-model-only scoring) evaluates trajectories solely using confidence scores from their source models.}
\label{tab:ab_cross_model}
\begin{tabular}{l cc cc}
\toprule
& \multicolumn{2}{c}{\textbf{General}} & \multicolumn{2}{c}{\textbf{Math}} \\
\cmidrule(lr){2-3}\cmidrule(lr){4-5}
\textbf{Method} & MMLU$^*$ & ARC-C & GSM8K & MATH500 \\
\midrule
TIE w/o cross-model scoring & 77.28 & 88.23 & 81.35 & 42.4 \\
TIE & \textbf{77.69} & \textbf{88.82} & \textbf{82.71} & \textbf{47.0} \\
\bottomrule
\end{tabular}%
\end{table*}

In this section, we discuss why cross-model scoring is important when using logit-based scoring functions (\ie top-1 confidence, entropy, and confidence margin). Since different models are calibrated differently, directly comparing their logits may lead to biased trajectory comparisons. For example, if one model tends to be overly confident in its answers, logit-based scoring functions may disproportionately favor its trajectories regardless of their actual reliability. Therefore, accounting for calibration differences across models is essential for robust trajectory comparison.

To address this, we employ cross-model scoring, which evaluates a given trajectory not only under its source model but across all constituent models, selecting the trajectory that exhibits the highest confidence on average. As shown in \Cref{tab:ab_cross_model}, cross-model scoring outperforms source-model-only scoring by favoring trajectories that are consistently supported by all constituent models. These results suggest that accounting for calibration differences is important when scoring trajectories using logit-based scoring functions.

\section{Ablation on Final Response Selection Strategies}
\label{sec:appendix_response_selection}

\begin{table*}[!t]
\centering
\small
\setlength{\tabcolsep}{4pt}
\caption{Ablation on the final response selection strategy. Token change count is used for trajectory assessment. We compare three strategies: (i) \emph{lowest TCC}, selecting the response with the lowest $\tilde{\mathcal{C}}_m^{(T)}$ (\Cref{subsec:method_4}); (ii) \emph{best model}, selecting the response from the best-performing individual model; and (iii) \emph{most selected}, selecting the response from the model chosen most frequently during ensembling.}
\label{tab:ab_response_selection}
\begin{tabular}{l cc cc c}
\toprule
 & \multicolumn{2}{c}{\textbf{General}} & \multicolumn{2}{c}{\textbf{Math}} & \textbf{Coding} \\
\cmidrule(lr){2-3}\cmidrule(lr){4-5}\cmidrule(lr){6-6}
\textbf{Method} & MMLU$^*$ & ARC-C & GSM8K & MATH500 & MBPP \\
\cmidrule[1.5pt]{1-6}
\multicolumn{4}{l}{\textit{Single Models}} \\
\cmidrule[1.5pt]{1-6}
LLaDA & 71.45 & 85.15 & 78.77 & 37.4  & 53.16  \\
Dream    & 77.37 & 86.69 & 78.39 & 48.0  & 63.23 \\
\cmidrule[1.5pt]{1-6}
\multicolumn{4}{l}{\textit{Intermediate-generation Ensemble}} \\
\cmidrule[1.5pt]{1-6}
(i) Lowest TCC
& 78.12 & \textbf{89.16} & \textbf{83.47} & 48.6 & 64.17 \\
(ii) Best model
& \textbf{78.64} & 88.23 & 81.65 & \textbf{49.0} & \textbf{66.04} \\
(iii) Most selected & 78.06 & 88.82 & 83.17 &  48.2 & 63.00 \\
\bottomrule
\end{tabular}%

\end{table*}

There is no trajectory relay after the final trajectory generation step since all answer-token positions have been unmasked. Therefore, an additional strategy is required to select the final response among the $M$ candidate responses. We explore three final response selection strategies: (i) \textit{lowest TCC}, which selects the response with the lowest $\tilde{\mathcal{C}}_m^{(T)}$; (ii) \textit{best model}, which selects the response from the best-performing individual model (Dream in our setup); and (iii) \textit{most selected}, which selects the response from the model that produced the highest-scoring trajectory most frequently during ensembling. As shown in \Cref{tab:ab_response_selection}, all three strategies achieve strong performance. However, in domains with a large performance gap between models, the \textit{best model} strategy performs particularly well. This indicates that when a clearly superior model exists, simply selecting its final response is effective. In contrast, when constituent models exhibit comparable performance, selecting the response with the lowest $\tilde{\mathcal{C}}_m^{(T)}$ produces better results. Overall, these results suggest that the optimal final response selection strategy may depend on the relative performance gap between constituent models.

\section{Anchor-free Variant of TIE}
\label{sec:appendix_anchor_free}

\begin{table}[!ht]
\centering
\small
\setlength{\tabcolsep}{4pt}
\caption{Results of the anchor-free variant of TIE when ensembling LLaDA and Dream. Both variants use token change count for trajectory assessment.}
\label{tab:ab_anchor_free}
\resizebox{\columnwidth}{!}{%
\begin{tabular}{l cc c c}
\toprule
 & \multicolumn{2}{c}{\textbf{General}} & \textbf{Math} & \textbf{Planning} \\
\cmidrule(lr){2-3}\cmidrule(lr){4-4}\cmidrule(lr){5-5}
\textbf{Method} & MMLU$^*$ & ARC-C & GSM8K & Countdown \\
\midrule
LLaDA & 71.45 & 85.15 & 78.77 & 13.4 \\
Dream & 77.37 & 86.69 & 78.39 & 16.4 \\
\midrule
TIE & \textbf{78.12} & \textbf{89.16} & \textbf{83.47} & 18.8 \\
TIE (anchor-free) & 78.08 & 88.40 & 83.32 & \textbf{19.8} \\
\bottomrule
\end{tabular}%
}
\vspace{-0.1in}
\end{table}

TIE relies on answer-token positions, which are defined by appending an answer suffix (\Cref{sec:appendix_answer_token_pos}). To examine whether TIE depends on this design, we evaluate an anchor-free variant in which no answer suffix is appended and no answer positions are defined. Instead, the anchor-free variant computes the token change count over the first four masked positions of the generation region at each decoding step, excluding positions whose top-1 token is the EOS token. Everything other than the token span used for the token change count remains unchanged.

As shown in \Cref{tab:ab_anchor_free}, the anchor-free variant achieves performance comparable to the original TIE, indicating that the effectiveness of TIE does not hinge on predefined answer positions.

\FloatBarrier
\section{Ensembling More Than Two Models}
\label{sec:appendix_three_models}

\begin{table}[!ht]
\centering
\small
\setlength{\tabcolsep}{4pt}
\caption{Results of ensembling three MDLMs with TIE on ARC-C.}
\label{tab:ab_three_models}
\begin{tabular}{l c}
\toprule
\textbf{Method} & ARC-C \\
\midrule
LLaDA & 85.15 \\
Dream & 86.69 \\
DreamCoder & 83.62 \\
\midrule
TIE (LLaDA + Dream) & 89.16 \\
TIE (LLaDA + Dream + DreamCoder) & \textbf{89.51} \\
\bottomrule
\end{tabular}
\vspace{-0.1in}
\end{table}

TIE is not limited to two constituent models: at each trajectory assessment step, the partial trajectories of all constituent models are scored, and the most reliable one is relayed to the others. To examine whether TIE benefits from additional models, we add DreamCoder to the ensemble of LLaDA and Dream and evaluate on ARC-C. As shown in \Cref{tab:ab_three_models}, the three-model ensemble further improves over the two-model ensemble, even though DreamCoder alone underperforms both LLaDA and Dream. This suggests that TIE can effectively leverage complementary knowledge from a larger pool of MDLMs.

\FloatBarrier
\section{Consistency of Gains and Statistical Reliability}
\label{sec:appendix_stat}

\begin{table}[!ht]
\centering
\small
\setlength{\tabcolsep}{4pt}
\caption{Results of ensembling LLaDA and Dream under stochastic decoding (temperature 0.2). We report the mean $\pm$ standard error over three random seeds.}
\label{tab:ab_seed}
\resizebox{\columnwidth}{!}{%
\begin{tabular}{l ccc}
\toprule
\textbf{Method} & ARC-C & WinoGrande & MBPP \\
\midrule
LLaDA & 85.10 {\scriptsize $\pm$ 0.27} & 70.93 {\scriptsize $\pm$ 0.39} & 52.85 {\scriptsize $\pm$ 0.77} \\
Dream & 87.09 {\scriptsize $\pm$ 0.22} & 71.27 {\scriptsize $\pm$ 0.16} & 63.78 {\scriptsize $\pm$ 0.08} \\
\midrule
TIE & \textbf{89.08} {\scriptsize $\pm$ 0.09} & \textbf{73.40} {\scriptsize $\pm$ 0.45} & \textbf{64.48} {\scriptsize $\pm$ 0.48} \\
\bottomrule
\end{tabular}%
}
\vspace{-0.1in}
\end{table}

\paragraph{When does TIE yield consistent gains?}
The gains of TIE are not uniform across settings, and this pattern follows the analysis in \Cref{subsec:main_analysis}-(ii). TIE yields clear gains when the constituent models have comparable and strong capabilities on the target task. For example, on GSM8K, LLaDA and Dream achieve similar performance (78.77 vs.\ 78.39), and TIE improves over the stronger model by 4.70\%p (83.47). Likewise, on ARC-C, where both models perform strongly (85.15 and 86.69), TIE improves over the stronger model by 2.47\%p (89.16). Conversely, the benefit diminishes when the capability gap between the models widens or their overall capability on the target task decreases. In such cases, the weaker model mostly produces low-quality trajectories and contributes little complementary information to the ensemble. These uneven gains are therefore not random but follow a predictable pattern determined by the relative and absolute capabilities of the constituent models, which provides a practical criterion for when to apply TIE.

\paragraph{Variance under stochastic decoding.}
Our main experiments use greedy decoding, so every reported result is deterministic without run-to-run variance. To further examine statistical reliability, we evaluate TIE under stochastic decoding with a temperature of 0.2 using three random seeds. As shown in \Cref{tab:ab_seed}, TIE consistently outperforms both individual models on all three benchmarks with small standard errors, confirming that its improvements are not an artifact of a particular decoding run.

\FloatBarrier
\section{Qualitative examples}
This section presents qualitative examples of TIE. \Cref{tab:appendix_qual1,tab:appendix_qual2} compare the responses generated by individual models with those produced by ensembling the same models via TIE.

\begin{table*}
    \centering
    \begin{tabularx}{\linewidth}{p{1.7cm}|X}
    \toprule
       Question  & A merchant wants to make a choice of purchase between 2 purchase plans: jewelry worth \$5,000 or electronic gadgets worth \$8,000. His financial advisor speculates that the jewelry market will go up 2.5\% while the electronic gadgets market will rise 1.2\% within the same month. If the merchant is looking to maximize profit at the end of this month by making a choice, how much profit would this be? \\
       \midrule
        Answer & 125  \\
        \midrule
        LLaDA & To determine the potential profit for the merchant, we need to calculate the future value of the jewelry and electronic gadgets after the market increases and then find the difference. 1. Calculate the future value of the jewelry: - Current value of jewelry: \$5,000 - Market increase: 2.5\% - Future value of jewelry: \$5,000 * (1 + 0.025) = \$5,000 * 1.025 = \$5,125 2. Calculate the future value of the electronic gadgets: - Current value of electronic gadgets: \$8,000 - Market increase: 1.2\% - Future value of electronic gadgets: \$8,000 * (1 + 0.012) = \$8,000 * 1.012 = \$8,096 3. Calculate the profit: - Profit: Future value of electronic gadgets - Future value of jewelry = \$8,096 - \$5,125 = \$71 Answer: \textcolor{blue}{$\boxed{71}$} \\
        \midrule
        Dream & 1. Calculate the potential increase for the jewelry market: - Initial value of jewelry: \$5,000 - Increase percentage: 2.5\% - Increase: \$5,000 * 0.025 = \$125 - New value: \$5,000 + \$125 = \$5,125 2. Calculate the potential increase for the electronic gadgets market: - Initial value of electronic gadgets: \$8,000 - Increase percentage: 1.2\% - Increase: \$8,000 * 0.012 = \$96 - New value: \$8,000 + \$96 = \$8,096 3. Compare the new values to determine which purchase plan yields a higher profit: - Jewelry market: \$5,125 - Electronic gadgets market: \$8,096 4. Calculate the profit for the electronic gadgets market: - Profit: New value - Cost = \$8,096 - \$8,000 = \$96 Answer: \textcolor{blue}{$\boxed{96}$} \\
        \midrule
        {\textbf{TIE} (LLaDA $+$ Dream)}
 &1. Calculate the potential increase for the jewelry market: - Initial value of jewelry: \$5,000 - Increase percentage: 2.5\% - Increase: \$5,000 * 0.025 = \$125 - New value: \$5,000 + \$125 = \$5,125 2. Calculate the potential increase for the electronic gadgets market: - Initial value of electronic gadgets: \$8,000 - Increase percentage: 1.2\% - Increase: \$8,000 * 0.012 = \$96  - New value: \$8,000 + \$96 = \$8,096 3. Determine the profit for each option: - Profit from jewelry: \$5,125 - \$5,000 = \$125 - Profit from electronic gadgets: \$8,096 - \$8,000 = \$96 4. Compare the profits: - Profit from jewelry is higher than electronic gadgets. Answer: \textcolor{red}{$\boxed{125}$}  \\
        \bottomrule
    \end{tabularx}
    \caption{Qualitative example of TIE. Both individual models reach incorrect answers (\textcolor{blue}{blue}): LLaDA makes a calculation error, and Dream selects the less profitable option. TIE retains the correct intermediate computations, arriving at the correct answer (\textcolor{red}{red}).}
    \label{tab:appendix_qual1}
\end{table*}

\begin{table*}
    \centering
    \begin{tabularx}{\linewidth}{p{1.7cm}|X}
    \toprule
       Question  & The vending machines sell chips for 40 cents and candy bars for 75 cents. George spent \$5 and got 3 bags of chips and had 1\% of his money left. How many candy bars did he buy? \\
       \midrule
        Answer & 5  \\
        \midrule
        LLaDA & 1. George spent \$5 in total. 2. He bought 3 bags of chips at 40 cents each, which is 3 * 40 = 120 cents. 3. He had 1\% of his money left, which is 1\% of \$5, or 0.01 * 500 = 5 cents. 4. Therefore, the total amount spent on chips and candy bars is 500 - 5 = 385 cents. 5. The amount spent on chips is 120 cents, so the amount spent on candy bars is 385 - 120 = 265 cents. 6. Each candy bar costs 75 cents, so the number of candy bars George bought is 265 / 75 = 3.5333... candy bars. 7. Since George can't buy a fraction of a candy bar, we need to find the whole number of candy bars he bought. 8. Rounding 3.5333... to the nearest whole number, we get that George bought 3 candy bars. Answer: \textcolor{blue}{$\boxed{3}$} \\
        \midrule
        Dream & 1. George spent \$5 and got 3 bags of chips. Each bag of chips costs 40 cents, so 3 bags of chips cost 3 * 40 = 120 cents. 2. George had 1\% of his money left after buying the chips. To find out how much money he had left, we need to find 1\% of \$5. 1\% of \$5 is 0.01 * 5 = 0.05 dollars. 3. George spent \$5 - 0.05 = \$4.95 on candy bars. 4. Each candy bar costs 75 cents. To find out how many candy bars George bought, we need to divide the amount he spent on candy bars by the cost of one candy bar. \$4.95 / 0.75 = 6.6. 5. Since George can't buy a fraction of a candy bar, we need to round down to the nearest whole number. So, George bought 6 candy bars. 6. However, we need to find the number of candy bars he bought, not the amount he spent. So, the answer is 6. Answer: \textcolor{blue}{$\boxed{6}$} \\
        \midrule
        {\textbf{TIE} (LLaDA $+$ Dream)}
 &1. George spent \$5 and got 3 bags of chips. Each bag of chips costs 40 cents, so the total cost for 3 bags is 3 * 40 = 120 cents. 2. George had 1\% of his money left. Since \$5 is equivalent to 500 cents, 1\% of 500 cents is 0.01 * 500 = 5 cents. So, George had 5 cents left. 3. The total amount George spent is \$5, which is equivalent to 500 cents. He spent 120 cents on chips and had 5 cents left, so the amount spent on candy bars is 500 - 120 - 5 = 375 cents. 4. Each candy bar costs 75 cents. To find out how many candy bars George bought, divide the total amount spent on candy bars by the cost of one candy bar: 375 / 75 = 5. 5. Therefore, George bought 5 candy bars. The final answer is 5, so the format of the final answer would be: Answer: \textcolor{red}{$\boxed{5}$}  \\
        \bottomrule
    \end{tabularx}
    \caption{Qualitative example of TIE. LLaDA makes an arithmetic error when computing the amount spent ($500 - 5 = 385$), and Dream ignores the cost of the chips when computing the amount spent on candy bars, both leading to incorrect answers (\textcolor{blue}{blue}). TIE correctly accounts for both the cost of the chips and the remaining money, and arrives at the correct answer (\textcolor{red}{red}).}
    \label{tab:appendix_qual2}
\end{table*}

\end{document}